\documentclass{article}
\usepackage{tikz-cd}
\usepackage{multirow}
\usepackage{booktabs}
\usepackage{amsmath}
\usepackage{float}
\usepackage{subfigure}
\usepackage{subcaption}
\usepackage{graphicx}
\usepackage{wrapfig}
\usepackage{amsfonts}
\usepackage{footmisc}
\usepackage[preprint]{corl_2024} 
\title{Equivariant Offline Reinforcement Learning}

\author{
  Arsh Tangri$^{*1}$,Ondrej Biza$^{1}$, Dian Wang$^{1}$, David Klee$^{1}$, Owen Howell$^{2}$, Robert Platt$^{1}$\\
  $^{1}$Khoury College of Computer Sciences, Northeastern University, Boston MA, 02115\\
  $^{2}$Department of Electrical and Computer Engineering, Northeastern University, Boston MA, 02115\\
}

\begin{document}
\maketitle


\begin{abstract}
    Sample efficiency is critical when applying learning-based methods to robotic manipulation due to the high cost of collecting expert demonstrations and the challenges of on-robot policy learning through online Reinforcement Learning (RL). Offline RL addresses this issue by enabling policy learning from an offline dataset collected using any behavioral policy, regardless of its quality. However, recent advancements in offline RL have predominantly focused on learning from large datasets. Given that many robotic manipulation tasks can be formulated as rotation-symmetric problems, we investigate the use of $SO(2)$-equivariant neural networks for offline RL with a limited number of demonstrations. Our experimental results show that equivariant versions of Conservative Q-Learning (CQL) and Implicit Q-Learning (IQL) outperform their non-equivariant counterparts. We provide empirical evidence demonstrating how equivariance improves offline learning algorithms in the low-data regime.

\end{abstract}


\renewcommand{\thefootnote}{\fnsymbol{footnote}}
\setcounter{footnote}{1}
\footnotetext{tangri.a@northeastern.edu}

\section{Introduction}
	


    Despite the recent breakthroughs in robot learning, learning optimal policies for robotic manipulation remains a challenging problem. Methods based on Imitation-Learning (IL) \cite{fang2019survey} have been shown to require a large number of expert demonstrations, which are difficult to collect due to the significant time and effort required. Conversely, reinforcement learning (RL) based methods learn policies through trial-and-error but require a large number of interactions with the environment to learn an optimal policy \cite{andrychowicz2020learning, kalashnikov2018scalable}. Offline RL \cite{levine2020offline} addresses these issues by enabling the learning of a policy without the explicit requirement of expert-demonstrations. It addresses the problem of learning a policy from a static dataset of transitions without making any assumptions about the behavioral policy used for collecting the transitions.
    
    Previous work in Offline RL \cite{kumar20conservative,kostrikov22offline,nair2020awac,peng2019advantage} has largely concentrated on developing algorithms capable of learning effective policies from large datasets and scaling efficiently with increasing dataset size. However, collecting even sub-optimal trajectories presents significant challenges due to episodic resets and hardware failures. Consequently, assembling large datasets may not always be practical, especially for sensitive or expensive robotics tasks. Therefore, focusing on learning from small, offline, sub-optimal datasets could be a more viable approach for learning robotics policies, though it remains a largely unsolved problem. 



    In this work, we investigate the effectiveness of Offline RL algorithms for learning robotic manipulation policies from limited datasets ranging from about 200 transitions up to 1000 transitions. Recent advancements have demonstrated the substantial benefits of utilizing equivariant neural networks in both online \cite{wang22so2} and offline \cite{huang2024leveraging,zhu2022sample} policy-learning settings due to their ability to reduce the amount of data required for learning optimal policies. Our work further showcases how techniques in geometric deep learning and Offline RL can be leveraged for robotic-manipulation tasks. Specifically, we introduce $SO(2)$-equivariant versions of Conservative Q-Learning (CQL, \cite{kumar20conservative}) and Implicit Q-Learning (IQL, \cite{kostrikov22offline}) for offline policy-learning from small datasets.

Our contributions are three-fold. First, we evaluate the efficacy of popular offline RL algorithms, CQL \cite{kumar20conservative} and IQL \cite{kostrikov22offline}, in learning robotic manipulation policies from small expert and suboptimal datasets. We find that these algorithms often struggle to outperform the behavioral policies used to collect the datasets. In order to remedy this, we introduce $SO(2)$-equivariant versions of CQL and IQL, which both outperform their non-equivariant counterparts and learn policies that \emph{exceed} the performance of the behavioral policies in the case of sub-optimal datasets. Third, we find that equivariance has a different effect on CQL and IQL due to their respective design principles. CQL's training includes an action sampling step that evaluates actions not represented in the dataset. Hence, CQL benefits from both an invariant critic, which mitigates extrapolation error of unseen actions, and an equivariant actor, which improves generalization at inference time. Conversely, IQL's critic never encounters actions outside of the dataset distribution and we find the invariant critic to be less important in this case. These results provide new insight into the role of generalization in the actor and the critic in Offline RL.

\section{Related Work}\label{RelatedWork}

    \textbf{Equivariant Deep Learning} pertains to encoding symmetries in the structure of neural networks \cite{cohen2016group, cohen2016steerable, weiler2019general, cesa2021program}. Recent works have demonstrated that Equivariant neural networks significantly improve the performance and sample efficiency in robotic manipulation tasks compared to their non-equivariant counterparts. 
    \citet{zhu2022sample,jia2023seil,wang2022robot} make use of $SO(2)$-equivariance to learn optimal manipulation and grasping policies with improved sample-efficiency. 
    Furthermore, \citet{huang2024leveraging} exploits the $SO(2)$ bi-equivariant symmetry of the pick-and-place problem to outperform prior similar methods. Several works have also attempted to leverage the $SO(3)$-symmetry in robotics. \citet{simeonov2022neural,ryu2022equivariant} encode $SO(3)$-symmetry by learning $SO(3)$-equivariant point features for solving 6-DoF robotic manipulation tasks. \citet{huang2024fourier} proposes a new transporter-like method for modelling $SO(3)$-bi equivariance. \citet{pan2023tax}  learns $SO(3)-$invariant point correspondences for pick-and-place tasks. In the context of Deep RL, equivariant neural networks have been shown to significantly improve sample-efficieny by \citep{vanderpol2021mdp,wang22so2,mondal2020group}. However, to the best of our knowledge, equivariant methods have not been explored for Offline RL.

    \textbf{Offline Reinforcement Learning} \cite{levine2020offline} relates to learning control policies from previously-collected offline datasets. Most of the methods are based on constrained Q-Learning, where constraints are implemented using divergence constrains \citet{nair2020awac}; \citet{peng2019advantage}, adding a behavior-cloning loss term \citet{fujimoto2021minimalist}, minimizing the Q-values for out-of-distribution actions \citet{kumar20conservative} or by avoiding out-of-distribution action sampling altogether \citet{kostrikov22offline}. Furthermore, works like \citet{chebotar2023q};\citet{kalashnikov2018scalable}; show the massive potential of Offline RL methods for scaling with increase dataset sizes and model sizes. However, not much attention has been paid  on learning from small datasets, which is what our work focuses on. 

\section{Background}

\textbf{Equivariance over $SO(2)$ Group:} Robotic manipulation problems inherently exhibit $SO(2)$-symmetry, encompassing the rotational symmetry in the plane perpendicular to gravity. The special orthogonal group $SO(2)$, which consists of all rotations about the origin, is defined as $SO(2) = {Rot_{\theta} : 0 \leq \theta < 2\pi}$. In this study, we focus on the cyclic group $C_n$, a discretized subset of $SO(2)$ generated by rotations of angle $\frac{2\pi}{n}$. Formally, the group $C_{n}$ is represented as
\begin{align}\label{Cyclic group definition}
C_n = \{ \text{Rot}_{\theta}: \theta \in \{ \frac{2\pi i}{n} | i\in\mathbb{Z}, 0 \leq i < n \} \}
\end{align}
Symmetries in problems can be described using the concepts of equivariant and invariant properties of functions. A function $f$ is said to be invariant under a group $G$ if the function's output remains unchanged when its input $x$ is transformed by any group element $g \in G$, formally expressed as $f(gx) = f(x)$. Similarly, a function is considered equivariant with respect to a group $G$ if transforming the input $x$ by $g \in G$ results in the output being transformed by the same group element, i.e., $f(gx) = gf(x)$.

\textbf{Group-Invariant MDPs and Equivariant Policy Learning}: Equivariant RL leverages the symmetries of the MDP to structure the neural networks used for representing the policy and the value functions \cite{wang22so2,vanderpol2021mdp}. An MDP $M = (S,A,R,T,\gamma)$ is said to be a group-invariant ($G-$invariant) MDP if $\forall g \in G$, $T(s,a,s') = T(gs,ga,gs')$ and $R(s,a) = R(gs,ga)$. G-invariant MDPs can effectively describe most robotic-manipulation problems. The optimal Q-function $Q^{*}$ for a $G$-invariant MDP is group invariant, $Q^{*}(gs,ga)=Q^{*}(s,a)$, while the optimal policy $\pi^{*}$ is group equivariant, $\pi^{*}(gs)=g\pi^{*}(s)$. Hence, equivariant Soft Actor-Critic (Equi-SAC) uses equivariant neural networks to encode the  symmetries of a $G$-invariant MDP. It models the critic as a $G$-invariant function, $Q(s,a) = Q(gs,ga)$, and models the actor as a $G$-equivariant function, $\pi(gs) = g\pi(s)$. 

\textbf{Offline Reinforcement Learning}:
In the Offline RL setting \cite{levine2020offline}, one has access to only a previous transcript of states, actions and rewards. Offline RL algorithms learn a policy that maximizes the expected cumulative return $J(\pi) = E_{s_{0} \sim p(s_{0}), a \sim \pi(a_{t}|s_{t}), s_{t+1} \sim T(s_{t+1}|s_{t},a_{t})}[\sum_{t=0}^{\infty}\gamma^{t} r(s_{t},a_{t})]$, entirely from a static dataset $D = \{(s,a,r,s')_i\}_{i=1}^{N}$ collected using some arbitrary behavioral policy $\pi_\beta$. The application of standard online policy improvement methods in the offline setting requires evaluating actions that are absent in the dataset, which then leads to erroneous overestimation Q-value estimations due to these actions being out-of-distribution \cite{kumar20conservative}. Offline RL algorithms address this problem in multiple ways, namely avoiding out-of-distribution querying of value-functions and explicitly minimizing the Q-values of out-of-distribution actions.  

Conservative Q-Learning (CQL, \cite{kumar20conservative}) solves the problem of overestimation by explicitly minimizing the Q-values of unseen actions. It does so by changing the Q-Learning objective in the following way:
\begin{align}\label{equation:Offline-RL CQL Objective}
    L_{CQL}(\phi) = {E}_{(s,a,s')\sim D}\Big[&{E}_{\mu(a | s)}[Q_{\phi}(s,a)] - E_{\pi_{\beta}(a | s)}[Q_{\phi}(s,a)] \, + \nonumber \\
    &(r(s,a) + \gamma {E}_{a'\sim \pi(a'|s')}[Q_{\hat\phi}(s',a')] - Q_{\phi}(s,a))^2\Big]
\end{align}
where $\pi_\beta$ is the behavioral policy used for collecting the dataset, and $\mu$ is a distribution defined over the action-space such that actions with high a Q-value are more likely to be sampled. The rest of the algorithm remains the same as SAC \cite{haarnoja18soft}.

Implicit Q-Learning (IQL, \cite{kostrikov22offline}), on the other hand, uses expectile-regression to approximate the expectile $\tau$ over the distribution of actions using following objectives:
\begin{align}
L_{V}(\psi) &= {E}_{(s,a)\sim D}[L_{2}^{\tau}(Q_{\hat\phi}(s,a) - V_\psi(s))] \label{equation:Offline-RL IQL V Objective} \\
L_Q(\phi) &= {E}_{(s,a,s')\sim D}[(r(s,a) + \gamma V_\psi(s') - Q_{\phi}(s,a))^2] \label{equation:Offline-RL IQL Q Objective}
\end{align}
where $L_{2}^{\tau}(u) = |\tau - \mathbb{I}(u < 0)|u^2$. The use of a high expectile allows us to learn optimal value functions without sampling actions from an explicit-policy, and only from the dataset. For policy extraction, we use the following objective:
\begin{equation}\label{equation:Offline-RL IQL Policy Objective}
L_{\pi}(\theta) = {E}_{(s,a)\sim D}[\exp(\alpha(Q_{\hat\phi}(s,a) - V_\psi(s)))\log \pi_{\theta}(a|s)]
\end{equation}

\section{Equivariant Offline Reinforcement Learning}
\subsection{$SO(2)$-Invariant MDPs}
In this work, we focus on the robotic manipulation tasks in the two-dimensional plane. This class of problems is described by $SO(2)$-invariant MDPs \cite{wang22so2,vanderpol2021mdp}. We use $C_8$ group \ref{Cyclic group definition} to approximate the $SO(2)$-group. The state of the MDP is expressed as an $m$-channel feature map, $f_s: R^2 \rightarrow R^m$. A group element $g \in C_8$ acts on the image in the following way: 
\begin{equation}\label{equation:TransformationImage}
(gf_s)(x,y) = \rho_{0}(g)f_s(\rho_{1}^{-1}(x,y))
\end{equation}
where $\rho_1$ symbolizes the rotation of pixel-locations, while $\rho_0$ indicates that the order of feature-channels remains unchanged. Furthermore, we assume that the MDP has a factored action-space $A_{inv} \times A_{equi} = A \subseteq R^k$. The elements in $A_{inv}$ are invariant to $g \in C_8$, whereas the elements in $A_{equi}$ transform with representation $\rho_{equi} = \rho_{1}$. Thus, $g \in C_8$ transforms $a = (a_{equi},a_{inv}), a_{equi} \in A_{equi}, a_{inv} \in A_{inv}$ in the following manner:
\begin{equation}\label{equation:TransformationAction}
ga = (\rho_{equi}(g)a_{equi}, a_{inv})
\end{equation}

\subsection{$SO(2)$-Invariant MDP for Robotic-Manipulation}
In this subsection, we describe the $SO(2)-$invariant MDP formulation for our robotic-manipulation tasks. A $2$-channel image is used to represent the state of the MDP, where the first channel is top-down depth image centered with respect to the robot gripper, and the second channel is a binary channel indicating whether the gripper is holding an object or not. The action can be described by a tuple, $a = (a_{\lambda}, a_{xy}, a_z, a_{\theta})$, where $a_{\lambda}$ denotes the commanded gripper state (open or closed), $a_{xy}$ is the relative change in the $xy$ position of the gripper, $a_z$ is the relative change in the $z$ position of the gripper, and the $a_{\theta}$ is the commanded change in gripper-orientation. Furthermore, $a_{xy}$ is equivariant to $g \in C_8$ $A_{equi} = A_{xy}$, whereas, the other variables are invariant to the group-action $A_{inv} = A_{z} \times A_{\theta} \times A_{\lambda}$.

\subsection{$SO(2)$-Equivariant CQL and IQL}

\textbf{Conservative Q-Learning (CQL)}, analogous to Soft Actor-Critic (SAC) \cite{haarnoja18soft} in the online RL context, necessitates the learning of parameters for two neural networks: the policy network $\pi_{\theta}$ and the state-action value function $Q_{\phi}(s,a)$. Given the similarities between CQL and SAC, we leverage the design principles proposed in \citet{wang22so2} to develop an equivariant variant of CQL. We first describe the equivariance encoded in the actor and then the invariance encoded in the critic.


The actor network $\pi_{\theta}: S \rightarrow A \times A_{\sigma}$ estimates a conditional isotropic Gaussian distribution over the action space by outputting the mean and standard deviation, thus defining $\overline{A} = A \times A_{\sigma}$. To specify the action of $g \in C_8$ on the output from the policy network, we assume that $a_{\sigma}$ is invariant under the group operator, which is reasonable since the variance of the policy should remain unaffected by the choice of frame. Therefore, $g \in C_8$ acts on $\overline{a} \in \overline{A}$ as follows:
\begin{equation}\label{equation:TransformationCombinedAction}
g\overline{a} = (\rho_{equi}(g)a_{equi}, a_{inv}, a_{\sigma})
\end{equation}
The state-action critic uses an equivariant state encoder for encoding the state onto a regular representation $\overline{s} = e(f_s) \in (R^n)^\alpha$, where $n$ is the number of group elements and $\alpha$ is the size of the group ($\alpha=8$ for $C_8$). $\overline{s}$ is then concatenated with the action $a \in A$, which is then mapped to a Q-value estimate for the $(s,a)$ pair. The critic satisfies the following invariance equation:
\begin{equation}\label{equation:TransformationQFunction}
Q_{\phi}(e(gf_s),ga) = Q_{\phi}(e(f_s),a).
\end{equation}
\textbf{Implicit Q-Learning (IQL)}, while using a different loss functions, employs a similar actor-critic architecture to CQL. One difference is that the actor only outputs the mean of the distribution, while learning the standard-deviation as an unconditioned learnable parameter. Hence, we do not output the standard-deviation, and the policy follows the following equivariance relation for action $a=(a_{equi},a_{inv}) \in A$: 
\begin{equation}\label{equation:TransformationSingleAction}
ga = (\rho_{equi}(g)a_{equi}, a_{inv}).
\end{equation}
We use the same architecture for the state-action critic as in Equi-CQL. For the state-value function, we use a very similar architecture to the state-action value critic, including the use of an encoder $e$. The state-value function has the following invariance relation:
\begin{equation}\label{equation:TransformationVFunction}
V_{\psi}(e(gf_s)) = V_{\psi}(e(f_s)).
\end{equation}
We provide further architectural details in the appendix.

\section{Experiments}

Our experiments are designed to address several critical questions. Firstly, we investigate whether existing Offline RL algorithms can learn effective robotic manipulation policies from small offline datasets. Secondly, we examine whether the incorporation of $SO(2)$-equivariant neural networks can enhance the performance of these algorithms. Lastly, we aim to identify which specific aspects of adding equivariance contribute to the improved performance of these algorithms.


\begin{figure}[t]
    \centering
    \subfigure[Drawer Opening]{%
        \includegraphics[width=0.15\textwidth]{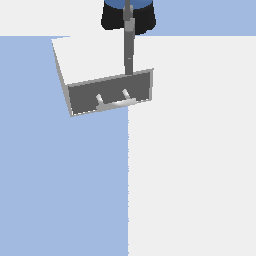}
        \includegraphics[width=0.15\textwidth]{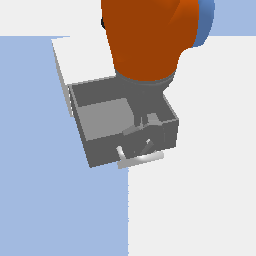}
        \label{fig:drawer_opening}
    }\hfill
    \subfigure[Block in Bowl]{%
        \includegraphics[width=0.15\textwidth]{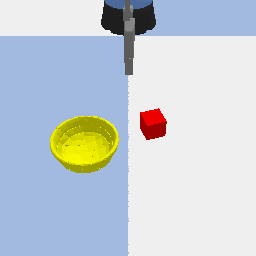}
        \includegraphics[width=0.15\textwidth]{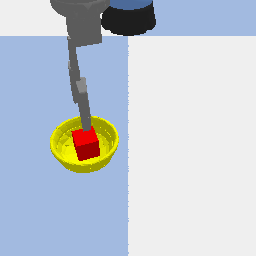}
        \label{fig:block_in_bowl}
    }\hfill
    \subfigure[Block Stacking]{%
        \includegraphics[width=0.15\textwidth]{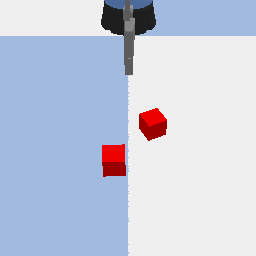}
        \includegraphics[width=0.15\textwidth]{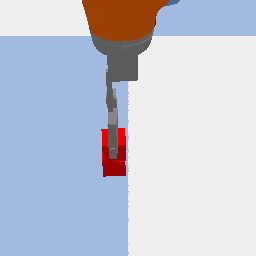}
        \label{fig:block_stacking}
    }
    \caption{Pybullet Tasks.  We show the state upon reset (left) and the respective goal state achieved by an expert (right). The position and rotation of the objects is randomized at each reset.}
    \label{fig:env_images}
    \vspace{0.2cm}
\end{figure}

\subsection{Manipulation Tasks}
We utilize BulletArm \cite{wang2022bulletarm}, a robotic manipulation benchmark built on PyBullet, to evaluate our methods on the tasks illustrated in the figure \ref{fig:env_images}. For all environments, the continuous action space is defined as the following: $A_\lambda = [0, 1]$ (Gripper) ; $ A_{xy} = \{(x,y) | x,y \in [-0.05m, 0.05m]\};A_z = [-0.05m, 0.05m]; A_{\theta}=[-\theta / 8, \theta / 8]$. All tasks have a binary reward function i.e +1 on task completion and 0 otherwise.

\subsection{Offline Datasets}
Our methods are tested on datasets comprising of all optimal transitions from an expert planner as well as datasets containing trajectories from a sub-optimal agent. The sub-optimal datasets include a mix of successful and failed episodes. We compare Equi-CQL and Equi-IQL against their non-equivariant versions. Each method is trained for a 100K gradient updates, with the exception of non-equivariant CQL, which exhibited significantly slower convergence and thus was trained for $300$K gradient updates for sub-optimal datasets and 200K gradient updates for the optimal-datasets. We also apply a rotational data augmentation to the transitions in each sampled batch for all methods, where a transition is augmented with a random $SO(2)$ rotation. We evaluated the methods every $1,000$ gradient updates using $100$ evaluation episodes and, similar to the approach in \cite{robomimic2021}, report the mean and standard deviation of the best evaluation discounted return for $3$ training seeds.

The sub-optimal datasets are created by training a policy using Equivariant SACfd until a certain pre-decided discounted return is reached, and then collecting a specific number of episodes or transitions using the sub-optimal policy. In our experiments, we set the number of episodes to $5$ and $10$. Furthermore, we also compare our algorithms on datasets with $1000$ transitions, which is roughly equivalent to 30 episodes for the tasks chosen. For collecting ``medium'' datasets, we set the discounted return limit to $0.4$ for early-stopping the policy-training, and $0.2$ for collecting ``near-random'' datasets.

The optimal datasets are collected using the expert-planner for each task provided by the BulletArm \cite{wang2022bulletarm} benchmark. We evaluate on datasets with 1, 5, and 10 expert demonstrations.

\subsection{Implementation Details}
For IQL, we determined that an expectile value of \(\tau=0.8\) and an inverse temperature of \(\beta=0.5\) yielded the best performance. In the case of CQL, we set the CQL-weight \(\alpha=1\) and CQL-temperature to 1. Additionally, we adopted the default hyperparameters from Equi-SAC, with the exceptions of a learning rate set to \(1 \times 10^{-4}\) and a network update rate of \(5 \times 10^{-3}\).
\subsection{Results}


Table \ref{table:OptimalDatasets} presents our results obtained on the optimal datasets, while Tables \ref{table: SubOptimalDatasets_5} and \ref{table: SubOptimalDatasets_10} display our findings on sub-optimal datasets with 5 and 10 demos respectively. Our results for the ablation study that focuses on elucidating the contributions of equivariance for these Offline RL algorithms are presented in Table \ref{table:  Ablation}. Finally, we benchmark our equivariant methods on sub-optimal datasets with 1000 transitions to examine if the benefits of equivariance persist with larger datasets, and the results are showcased in Table \ref{table: SubOptimalDatasets_1000} .

\begin{table}[t]
    \centering
    \small
    \centerline{
    \begin{tabular}{lc|cccc}
        \toprule
        \textbf{Task} & \textbf{Num. Demos} &  \textbf{Equi CQL} &  \textbf{Equi IQL} &  \textbf{Non-Equi CQL} & \textbf{Non-Equi IQL}\\
        \midrule
        \multirow{3}{*}{Drawer-Opening} & 1 & 0.4220\scriptsize{$\pm$0.1769} & \textbf{0.4654\scriptsize{$\pm$0.2007}} & 0.3108\scriptsize{$\pm$0.0451}&0.2479\scriptsize{$\pm$0.0439} \\ 
        & 5 &  0.6937\scriptsize{$\pm$0.0356} & \textbf{0.8182\scriptsize{$\pm$0.0390}} & 0.4147\scriptsize{$\pm$0.0787} & 0.7013\scriptsize{$\pm$0.0613} \\
        & 10 &  0.6792\scriptsize{$\pm$0.06518} & \textbf{0.8861\scriptsize{$\pm$0.0054}} & 0.5970\scriptsize{$\pm$0.0505} & 0.8408\scriptsize{$\pm$0.0204} \\ 
        \midrule
        \multirow{3}{*}{Block-in-Bowl} & 1 & \textbf{0.3156\scriptsize{$\pm$0.1574}} & 0.2510\scriptsize{$\pm$0.2045} & 0.2190\scriptsize{$\pm$0.1571} & 0.2385\scriptsize{$\pm$0.1163}\\
        & 5 & \textbf{0.7769\scriptsize{$\pm$0.0282}} & 0.6446\scriptsize{$\pm$0.0623} &0.0\scriptsize{$\pm$0.0}& 0.3796\scriptsize{$\pm$0.0669} \\
        & 10 & \textbf{0.7822\scriptsize{$\pm$0.0174}} &0.6458\scriptsize{$\pm$0.1175}& 0.0100\scriptsize{$\pm$0.0031} & 0.5937\scriptsize{$\pm$0.0938}\\
        \midrule
        \multirow{3}{*}{Block-Stacking} & 1 &\textbf{0.0865\scriptsize{$\pm$0.0430}}& 0.0660\scriptsize{$\pm$0.0378} & 0.0115\scriptsize{$\pm$0.0106} & 0.0713\scriptsize{$\pm$0.0478}\\
        & 5 & \textbf{0.4648\scriptsize{$\pm$0.0734}} & 0.3685\scriptsize{$\pm$0.0165} &0.0285\scriptsize{$\pm$0.0202} &0.2613\scriptsize{$\pm$0.0650} \\
        & 10 & 	0.4883\scriptsize{$\pm$0.0636} & 0.4515\scriptsize{$\pm$0.0359} &0.046\scriptsize{$\pm$0.0178}& \textbf{0.5017\scriptsize{$\pm$0.0333}} \\
        \bottomrule
    \end{tabular}
    }
    \caption{Best Evaluation Discounted Returns on Optimal Datasets, averaged over 3 seeds (Mean and Std Deviation)}
    \label{table:OptimalDatasets}
\end{table}

\subsubsection{Non-Equivariant Offline RL algorithms}
We find that that IQL demonstrates reasonable performance with these smaller datasets, successfully learning policies that are close to or slightly better than the behavioral policy in the sub-optimal case. CQL, on the other hand, often struggles to learn an effective policy from small offline datasets, despite being trained for a greater number of gradient updates. This issue is particularly pronounced in long-horizon tasks like Block-in-Bowl. The performance gap between IQL and CQL can be traced to the critic's limited generalization capacity in CQL, which struggles to accurately evaluate out-of-distribution actions in the TD-update and actor loss. This challenge is further compounded by the scarcity of data. In contrast, IQL avoids this issue entirely by not involving out-of-distribution action-sampling, thereby maintaining robust performance even with limited datasets.
\begin{wrapfigure}[11]{r}{0.3\textwidth}
  \vspace{-0.4cm}
  \centering
  \includegraphics[width=1.05\linewidth]{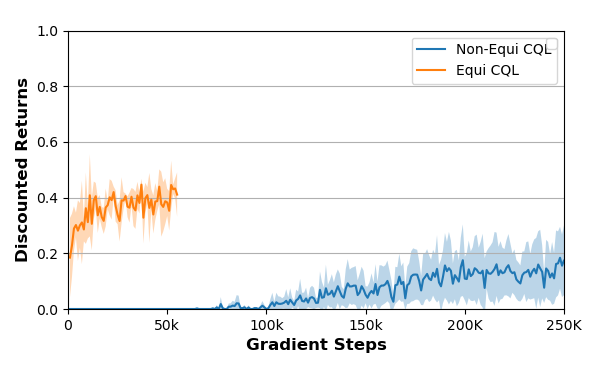}
  \caption{Performance of Equi-CQL, Non-Equi CQL as a function of gradient steps, on the Block-in-Bowl-Medium dataset with 10 episodes}
  \label{fig:helpful-occlusion}
\end{wrapfigure}
\subsubsection{Equivariant Offline RL algorithms}
Crucially, we find that Equi-CQL and Equi-IQL consistently outperform their non-equivariant counterparts across nearly all tasks. Specifically, Equi-IQL achieves superior performance with sub-optimal datasets, while Equi-CQL excels with optimal datasets. To elucidate the contributions of equivariance for these algorithms, we conducted ablation studies pairing the equivariant actor with a non-invariant critic, and conversely, a non-equivariant actor with an invariant critic. 

As seen in Table \ref{table: Ablation}, IQL with an equivariant actor and a non-invariant critic performs comparably to fully-equivariant IQL, whereas IQL with a non-equivariant actor and an invariant critic performs worse, akin to fully non-equivariant IQL. Hence, the performance improvement in Equi-IQL over non-equivariant IQL can be attributed primarily to its equivariant actor, which aids in policy generalization to unseen states. The invariant critic and value function, on the other hand, do not contribute to this improvement due to the lack of out-of-distribution action-sampling during training, rendering the improved generalization of an $SO(2)$-invariant critic and value function unnecessary for the accurate evaluation of unseen states and actions. 

\begin{table}[t]
    \centering
    \small
    \renewcommand{\arraystretch}{1.5} 
    \centerline{
    \begin{tabular}{ll|ccccc}
        \toprule
        \textbf{Task} & \textbf{Dataset Type} &  \textbf{Equi CQL} &  \textbf{Equi IQL} &  \textbf{Non-Equi CQL} & \textbf{Non-Equi IQL}\\
        \midrule
        \multirow{2}{*}{Drawer-Opening} & Near-Random &\textbf{0.3263\scriptsize{$\pm$0.04933}}&0.1894\scriptsize{$\pm$0.0880}&0.228\scriptsize{$\pm$0.114}
&0.1502\scriptsize{$\pm$0.1000}\\ 
        & Medium &\textbf{0.4473\scriptsize{$\pm$0.0792}} &0.3499\scriptsize{$\pm$0.023}&0.3034\scriptsize{$\pm$0.0586}&0.2628\scriptsize{$\pm$0.0817}\\
        \midrule
        \multirow{2}{*}{Block-in-Bowl} & Near-Random &0.2036\scriptsize{$\pm$0.1582}&\textbf{0.3837\scriptsize{$\pm$0.3006}}&0.1804\scriptsize{$\pm$0.1832}&0.3459\scriptsize{$\pm$0.2732}\\ 
        & Medium & \textbf{0.4985\scriptsize{$\pm$0.0547}} & 0.4828\scriptsize{$\pm$0.0808} & 0.2898\scriptsize{$\pm$0.1508}&0.4054\scriptsize{$\pm$0.0113}\\
        \midrule
        \multirow{2}{*}{Block-Stacking} & Near-Random &0.1131\scriptsize{$\pm$0.0523} & \textbf{0.2238\scriptsize{$\pm$0.0370}} & 0.0411\scriptsize{$\pm$0.0291}&0.1844\scriptsize{$\pm$0.0559}\\ 
        & Medium & 0.2422\scriptsize{$\pm$0.1653} & 0.4998\scriptsize{$\pm$0.0928} &0.3631\scriptsize{$\pm$0.0953}& \textbf{0.5432\scriptsize{$\pm$0.0544}} \\
        \bottomrule
    \end{tabular}
    }
\caption{Best Evaluation Discounted Returns on Sub-Optimal Datasets with 5 episodes, averaged over 3 seeds (Mean and Std Deviation)}
\label{table: SubOptimalDatasets_5}
\end{table}

\begin{table}[H]
    \centering
    \small
    \renewcommand{\arraystretch}{1.5} 
    \centerline{
    \begin{tabular}{ll|ccccc}
        \toprule
        \textbf{Task} & \textbf{Dataset Type} &  \textbf{Equi CQL} &  \textbf{Equi IQL} &  \textbf{Non-Equi CQL} & \textbf{Non-Equi IQL}\\
        \midrule
        \multirow{2}{*}{Drawer-Opening} & Near-Random &\textbf{0.3101\scriptsize{$\pm$0.1283}}&\textbf{0.3106\scriptsize{$\pm$0.1371}}&0.2741\scriptsize{$\pm$0.0783}&0.2934\scriptsize{$\pm$0.0328}\\ 
        & Medium &\textbf{0.5571\scriptsize{$\pm$0.0172}}&0.4601\scriptsize{$\pm $0.0748}&0.3160\scriptsize{$\pm$0.005} &0.4243\scriptsize{$\pm$0.0514}\\
        \midrule
        \multirow{2}{*}{Block-in-Bowl} & Near-Random &0.2475\scriptsize{$\pm$0.1542} & \textbf{0.4021\scriptsize{$\pm$0.0419}} &0.0027\scriptsize{$\pm$0.0038}& \textbf{0.4065\scriptsize{$\pm$0.0436}}\\ 
        & Medium & \textbf{0.6021\scriptsize{$\pm$0.0583}} & 0.5743\scriptsize{$\pm$0.0728} &0.0423\scriptsize{$\pm$0.0301}&0.5498\scriptsize{$\pm$0.0819} \\
        \midrule
        \multirow{2}{*}{Block-Stacking} & Near-Random & 0.3325\scriptsize{$\pm$0.0275} & \textbf{0.4165\scriptsize{$\pm$0.0531}} & 0.0046\scriptsize{$\pm$0.0034}  &0.3952\scriptsize{$\pm$0.0687}\\  
        & Medium &0.3986\scriptsize{$\pm$0.1793}& \textbf{0.4423\scriptsize{$\pm$0.0884}} &0.4235\scriptsize{$\pm$0.1326}&0.3854\scriptsize{$\pm$0.1561}\\
        \bottomrule
    \end{tabular}
    }
    \caption{Best Evaluation Discounted Returns on Sub-Optimal Datasets with 10 episodes, averaged over 3 seeds (Mean and Std Deviation)}
    \label{table: SubOptimalDatasets_10}
\end{table}


\vspace{-50pt} For CQL, we observe notable improvements over the non-equivariant version in both configurations: employing an equivariant actor with a non-invariant critic, and using an invariant critic with a non-equivariant actor. The enhancement provided by the equivariant actor can be attributed to its capacity to generalize to unseen states,similar to the benefits observed with Equi-IQL. Additionally, the invariant critic plays a crucial role in accurately evaluating out-of-distribution actions in CQL by generalizing from rotated versions of these actions present in the dataset, thereby enabling the learning of a more effective policy. Additionally, accurate evaluation of actions contributes to quicker convergence of Equi-CQL, as demonstrated in Fig.\ref{fig:helpful-occlusion}.



An \(SO(2)\)-invariant Q-function offers several advantages, including more effective minimization of Q-values for unseen actions. Specifically, minimizing the Q-value for a given \((s,a)\) inherently minimizes the Q-values for all corresponding \((g.s,g.a)\) across \(g \in G=C_8\). This property also mitigates CQL's overly conservative behavior by ensuring that all \((g.s,g.a)\) pairs, even those absent in the offline dataset, are assigned consistent Q-values. For instance, if a specific \((s,a)\) in the dataset reflects optimal behavior, the non-equivariant CQL might learn a high Q-value for this pair while minimizing Q-values for \((g.s,g.a)\) pairs not present in the dataset, despite these pairs also representing optimal behavior in rotated scenarios. Additionally, if the behavioral policy executes sub-optimal actions in such \(g.s\) states, non-equivariant CQL would erroneously elevate the Q-values for these actions, thus learning sub-optimal behavior. Conversely, the equivariant version maintains high Q-values across all \((g.s,g.a)\) pairs due to its invariance, thereby promoting optimal behavior across various \(g.s\) states. These points explain the superior performance of Equi-CQL over Non-Equi CQL, even with larger datasets. As illustrated in Table \ref{table: SubOptimalDatasets_1000}, the performance gap between Equi-IQL and Non-Equi IQL narrows due to a possible larger state-space coverage, thereby rendering the increased generalization of the actor inconsequential. However, this performance gap remains significant in the case of CQL.

\begin{table}[t]
    \centering
    \small
    \centerline{
    \begin{tabular}{ll|ccccc}
        \toprule
        \textbf{Dataset Type} & \textbf{Method} & \multicolumn{2}{c}{\textbf{Drawer Opening}} & \multicolumn{2}{c}{\textbf{Block in Bowl}} \\
        \cmidrule(r){3-4} \cmidrule(r){5-6}
        & & 5 & 10  & 5 & 10 \\
        \midrule
        \multirow{4}{*}{Near-Random} & Equi-CQL & 0.3263\scriptsize{$\pm$0.04933} & 0.3101\scriptsize{$\pm$0.1283} & 0.2036\scriptsize{$\pm$0.1582}  & \textbf{0.2475\scriptsize{$\pm$0.1542}}  \\
         &  w/ Non-Equi Actor& 0.2209\scriptsize{$\pm$0.1037} & 0.2563\scriptsize{$\pm$0.1815} & \textbf{0.2466\scriptsize{$\pm$0.1914}} & 0.0885\scriptsize{$\pm$0.0555}  \\
        &  w/ Non-Invar Critic  &  \textbf{0.3416\scriptsize{$\pm$0.0531}} & \textbf{0.3295\scriptsize{$\pm$0.0389}}  & 0.1479\scriptsize{$\pm$0.1479}  & 0.0056\scriptsize{$\pm$0.0079}   \\
        & Non-Equi CQL & 0.2280\scriptsize{$\pm$0.114} & 0.2741\scriptsize{$\pm$0.0783} & 0.1804\scriptsize{$\pm$0.1832}  & 0.0027\scriptsize{$\pm$0.0038}  \\
        \midrule
        \multirow{4}{*}{Medium} & Equi-CQL & \textbf{0.4473\scriptsize{$\pm$0.0792}} & \textbf{0.5571\scriptsize{$\pm$0.0172}} & \textbf{0.4985\scriptsize{$\pm$0.0547}}   & \textbf{0.6021\scriptsize{$\pm$0.0583}}  \\
         &  w/ Non-Equi Actor & 0.2968\scriptsize{$\pm$0.0455} & 0.3783\scriptsize{$\pm$0.1123} & 0.3884\scriptsize{$\pm$0.0313}  & 0.5438\scriptsize{$\pm$0.0096}  \\
         &  w/ Non-Invar Critic & 0.4277\scriptsize{$\pm$0.1356}
 & 0.3065\scriptsize{$\pm$0.0776}  & 0.3266\scriptsize{$\pm$0.0898}  & 0.2162\scriptsize{$\pm$0.1677}  \\
         & Non-Equi CQL   & 0.3034\scriptsize{$\pm$0.0586} & 0.3160\scriptsize{$\pm$0.005} & 0.2898\scriptsize{$\pm$0.1508}  & 0.0056\scriptsize{$\pm$0.0079}   \\
        \midrule
        \multirow{4}{*}{Near-Random}  & Equi-IQL & 0.1894\scriptsize{$\pm$0.0880} & \textbf{0.3106\scriptsize{$\pm$0.1371}} & \textbf{0.3837\scriptsize{$\pm$0.3006}}  & 0.4021\scriptsize{$\pm$0.0419} \\
        &  w/ Non-Equi Actor  & 0.1661\scriptsize{$\pm$0.1085} & 0.3191\scriptsize{$\pm$0.0393} & 0.2848\scriptsize{$\pm$0.2658}   & 0.3817\scriptsize{$\pm$0.0584}
  \\
        &  w/ Non-Invar Critic  & \textbf{0.2321\scriptsize{$\pm$0.0556}} & 0.3214\scriptsize{$\pm$0.1388}   & 0.3336\scriptsize{$\pm$0.2879}  & \textbf{0.4161\scriptsize{$\pm$0.0437}}  \\
        &  Non-Equi IQL & 0.1502\scriptsize{$\pm$0.1000} & 0.2934\scriptsize{$\pm$0.0328} & 0.3459\scriptsize{$\pm$0.2732}   & 0.4065\scriptsize{$\pm$0.0436}  \\
        \midrule
        \multirow{4}{*}{Medium} & Equi-IQL & 0.3499\scriptsize{$\pm$0.0230} & 0.4601\scriptsize{$\pm$0.0748} & 0.4828\scriptsize{$\pm$0.0808}  & 0.5743\scriptsize{$\pm$0.0728}  \\
        &  w/ Non-Equi Actor & 0.2305\scriptsize{$\pm$0.0384} & 0.3836\scriptsize{$\pm$0.03708} & 0.4257\scriptsize{$\pm$0.0470}  & 0.5402\scriptsize{$\pm$0.0433}  \\
        &  w/ Non-Invar Critic & \textbf{0.3594\scriptsize{$\pm$0.0285}} & \textbf{0.4875\scriptsize{$\pm$0.0529}} & \textbf{0.5002\scriptsize{$\pm$0.0522}} & \textbf{0.5933\scriptsize{$\pm$0.0171}}  \\
         & Non-Equi IQL & 0.2628\scriptsize{$\pm$0.0817} & 0.4243\scriptsize{$\pm$0.0514} & 0.4054\scriptsize{$\pm$0.0113}  & 0.5498\scriptsize{$\pm$0.0819}  \\
        \bottomrule
    \end{tabular}
    }
    \caption{Ablation Study: We evaluate Equi-IQL and Equi-CQL with 2 configurations, Equivariant Actor with a Non-Invariant Critic, and an Invariant Critic with a Non-Equivariant Actor, and compare the performance with their fully-equivariant and non-equivariant versions}
    \label{table:  Ablation}
\end{table}

\begin{table}[H]
    \centering
    \small
    \renewcommand{\arraystretch}{1.5} 
    \centerline{
    \begin{tabular}{ll|ccccc}
        \toprule
        \textbf{Task} & \textbf{Dataset Type} &  \textbf{Equi CQL} &  \textbf{Equi IQL} &  \textbf{Non-Equi CQL} & \textbf{Non-Equi IQL}\\
        \toprule
        \multirow{2}{*}{Drawer-Opening} & Near-Random &\textbf{0.4943\scriptsize{$\pm$0.1313}}&0.2406\scriptsize{$\pm$0.2004}&0.2764\scriptsize{$\pm$0.0754}& 0.2904\scriptsize{$\pm$0.0881}\\ 
        & Medium &0.5572\scriptsize{$\pm$0.0348} &0.5903\scriptsize{$\pm$0.0045}&0.4049\scriptsize{$\pm$0.0285}&\textbf{0.6066\scriptsize{$\pm$0.0387}}\\
        \midrule
        \multirow{2}{*}{Block-in-Bowl} & Near-Random &\textbf{0.5347\scriptsize{$\pm$0.0123}}&0.3711\scriptsize{$\pm$0.0934}&0.1489\scriptsize{$\pm$0.1459}&0.3935\scriptsize{$\pm$0.1816}\\ 
        & Medium &0.5806\scriptsize{$\pm$0.0364}&\textbf{0.6305\scriptsize{$\pm$0.0456}}&0.1961\scriptsize{$\pm$0.1768}&\textbf{0.6345\scriptsize{$\pm$0.0267}}\\
        \midrule
        \multirow{2}{*}{Block-Stacking} & Near-Random &0.3086\scriptsize{$\pm$0.0143}& \textbf{0.3489\scriptsize{$\pm$0.0012}}&0.1820\scriptsize{$\pm$0.0881}&0.3118\scriptsize{$\pm$0.0228}\\ 
        & Medium&0.4465\scriptsize{$\pm$0.0958}& \textbf{0.4986\scriptsize{$\pm$0.0846}} &0.2561\scriptsize{$\pm$0.2412}&0.5406\scriptsize{$\pm$0.0367}\\
        \bottomrule
    \end{tabular}
    }
\vspace{1pt}
\caption{Best Evaluation Discounted Returns on Sub-Optimal Datasets with 1000 transitions, averaged over 2 seeds. }
\label{table: SubOptimalDatasets_1000}
\end{table}


\section{Conclusion and Limitations}
\label{sec:conclusion}

In this work, we demonstrate that the performance of two prevalent offline RL methods for robotic manipulation tasks can be significantly enhanced by integrating an equivariant structure, especially when learning from small datasets. The challenge of collecting large, high-quality datasets remains a significant barrier in robotics, and our method addresses this issue by enabling the effective learning of policies from small and sub-optimal datasets. Our findings underscore the potential of symmetry-aware architectures to drive advancements in robot-learning, while also highlighting how equivariance can augments these algorithms.

\textbf{Limitations}: A natural limitation of our method is the assumption that a given problem is formulated as an $SO(2)$-invariant MDP. However, recent research \cite{wang2022surprising} has demonstrated that imposing equivariance constraints, even if they do not perfectly match the domain symmetry, can still result in performance improvements. We anticipate that our proposed methods will similarly benefit from this approach. 






\bibliography{example}  

\begin{thebibliography}{30}
\providecommand{\natexlab}[1]{#1}
\providecommand{\url}[1]{\texttt{#1}}
\expandafter\ifx\csname urlstyle\endcsname\relax
  \providecommand{\doi}[1]{doi: #1}\else
  \providecommand{\doi}{doi: \begingroup \urlstyle{rm}\Url}\fi

\bibitem[Fang et~al.(2019)Fang, Jia, Guo, Xu, Wen, and Sun]{fang2019survey}
B.~Fang, S.~Jia, D.~Guo, M.~Xu, S.~Wen, and F.~Sun.
\newblock Survey of imitation learning for robotic manipulation.
\newblock \emph{International Journal of Intelligent Robotics and Applications}, 3:\penalty0 362--369, 2019.

\bibitem[Andrychowicz et~al.(2020)Andrychowicz, Baker, Chociej, Jozefowicz, McGrew, Pachocki, Petron, Plappert, Powell, Ray, et~al.]{andrychowicz2020learning}
O.~M. Andrychowicz, B.~Baker, M.~Chociej, R.~Jozefowicz, B.~McGrew, J.~Pachocki, A.~Petron, M.~Plappert, G.~Powell, A.~Ray, et~al.
\newblock Learning dexterous in-hand manipulation.
\newblock \emph{The International Journal of Robotics Research}, 39\penalty0 (1):\penalty0 3--20, 2020.

\bibitem[Kalashnikov et~al.(2018)Kalashnikov, Irpan, Pastor, Ibarz, Herzog, Jang, Quillen, Holly, Kalakrishnan, Vanhoucke, et~al.]{kalashnikov2018scalable}
D.~Kalashnikov, A.~Irpan, P.~Pastor, J.~Ibarz, A.~Herzog, E.~Jang, D.~Quillen, E.~Holly, M.~Kalakrishnan, V.~Vanhoucke, et~al.
\newblock Scalable deep reinforcement learning for vision-based robotic manipulation.
\newblock In \emph{Conference on robot learning}, pages 651--673. PMLR, 2018.

\bibitem[Levine et~al.(2020)Levine, Kumar, Tucker, and Fu]{levine2020offline}
S.~Levine, A.~Kumar, G.~Tucker, and J.~Fu.
\newblock Offline reinforcement learning: Tutorial, review, and perspectives on open problems, 2020.

\bibitem[Kumar et~al.(2020)Kumar, Zhou, Tucker, and Levine]{kumar20conservative}
A.~Kumar, A.~Zhou, G.~Tucker, and S.~Levine.
\newblock Conservative q-learning for offline reinforcement learning.
\newblock In H.~Larochelle, M.~Ranzato, R.~Hadsell, M.~Balcan, and H.~Lin, editors, \emph{Advances in Neural Information Processing Systems 33: Annual Conference on Neural Information Processing Systems 2020, NeurIPS 2020, December 6-12, 2020, virtual}, 2020.
\newblock URL \url{https://proceedings.neurips.cc/paper/2020/hash/0d2b2061826a5df3221116a5085a6052-Abstract.html}.

\bibitem[Kostrikov et~al.(2022)Kostrikov, Nair, and Levine]{kostrikov22offline}
I.~Kostrikov, A.~Nair, and S.~Levine.
\newblock Offline reinforcement learning with implicit q-learning.
\newblock In \emph{The Tenth International Conference on Learning Representations, {ICLR} 2022, Virtual Event, April 25-29, 2022}. OpenReview.net, 2022.

\bibitem[Nair et~al.(2020)Nair, Gupta, Dalal, and Levine]{nair2020awac}
A.~Nair, A.~Gupta, M.~Dalal, and S.~Levine.
\newblock Awac: Accelerating online reinforcement learning with offline datasets.
\newblock \emph{arXiv preprint arXiv:2006.09359}, 2020.

\bibitem[Peng et~al.(2019)Peng, Kumar, Zhang, and Levine]{peng2019advantage}
X.~B. Peng, A.~Kumar, G.~Zhang, and S.~Levine.
\newblock Advantage-weighted regression: Simple and scalable off-policy reinforcement learning.
\newblock \emph{arXiv preprint arXiv:1910.00177}, 2019.

\bibitem[Wang et~al.(2022)Wang, Walters, and Platt]{wang22so2}
D.~Wang, R.~Walters, and R.~Platt.
\newblock {\textdollar}{\textbackslash}mathrm\{SO\}(2){\textdollar}-equivariant reinforcement learning.
\newblock In \emph{The Tenth International Conference on Learning Representations, {ICLR} 2022, Virtual Event, April 25-29, 2022}. OpenReview.net, 2022.

\bibitem[Huang et~al.(2024)Huang, Wang, Tangri, Walters, and Platt]{huang2024leveraging}
H.~Huang, D.~Wang, A.~Tangri, R.~Walters, and R.~Platt.
\newblock Leveraging symmetries in pick and place.
\newblock \emph{The International Journal of Robotics Research}, page 02783649231225775, 2024.

\bibitem[Zhu et~al.(2022)Zhu, Wang, Biza, Su, Walters, and Platt]{zhu2022sample}
X.~Zhu, D.~Wang, O.~Biza, G.~Su, R.~Walters, and R.~Platt.
\newblock Sample efficient grasp learning using equivariant models.
\newblock \emph{arXiv preprint arXiv:2202.09468}, 2022.

\bibitem[Cohen and Welling(2016{\natexlab{a}})]{cohen2016group}
T.~Cohen and M.~Welling.
\newblock Group equivariant convolutional networks.
\newblock In \emph{International conference on machine learning}, pages 2990--2999. PMLR, 2016{\natexlab{a}}.

\bibitem[Cohen and Welling(2016{\natexlab{b}})]{cohen2016steerable}
T.~S. Cohen and M.~Welling.
\newblock Steerable cnns.
\newblock \emph{arXiv preprint arXiv:1612.08498}, 2016{\natexlab{b}}.

\bibitem[Weiler and Cesa(2019)]{weiler2019general}
M.~Weiler and G.~Cesa.
\newblock General e (2)-equivariant steerable cnns.
\newblock \emph{Advances in neural information processing systems}, 32, 2019.

\bibitem[Cesa et~al.(2021)Cesa, Lang, and Weiler]{cesa2021program}
G.~Cesa, L.~Lang, and M.~Weiler.
\newblock A program to build e (n)-equivariant steerable cnns.
\newblock In \emph{International conference on learning representations}, 2021.

\bibitem[Jia et~al.(2023)Jia, Wang, Su, Klee, Zhu, Walters, and Platt]{jia2023seil}
M.~Jia, D.~Wang, G.~Su, D.~Klee, X.~Zhu, R.~Walters, and R.~Platt.
\newblock Seil: Simulation-augmented equivariant imitation learning.
\newblock In \emph{2023 IEEE International Conference on Robotics and Automation (ICRA)}, pages 1845--1851. IEEE, 2023.

\bibitem[Wang et~al.(2022)Wang, Jia, Zhu, Walters, and Platt]{wang2022robot}
D.~Wang, M.~Jia, X.~Zhu, R.~Walters, and R.~Platt.
\newblock On-robot learning with equivariant models.
\newblock \emph{arXiv preprint arXiv:2203.04923}, 2022.

\bibitem[Simeonov et~al.(2022)Simeonov, Du, Tagliasacchi, Tenenbaum, Rodriguez, Agrawal, and Sitzmann]{simeonov2022neural}
A.~Simeonov, Y.~Du, A.~Tagliasacchi, J.~B. Tenenbaum, A.~Rodriguez, P.~Agrawal, and V.~Sitzmann.
\newblock Neural descriptor fields: Se (3)-equivariant object representations for manipulation.
\newblock In \emph{2022 International Conference on Robotics and Automation (ICRA)}, pages 6394--6400. IEEE, 2022.

\bibitem[Ryu et~al.(2022)Ryu, Lee, Lee, and Choi]{ryu2022equivariant}
H.~Ryu, H.-i. Lee, J.-H. Lee, and J.~Choi.
\newblock Equivariant descriptor fields: Se (3)-equivariant energy-based models for end-to-end visual robotic manipulation learning.
\newblock \emph{arXiv preprint arXiv:2206.08321}, 2022.

\bibitem[Huang et~al.(2024)Huang, Howell, Zhu, Wang, Walters, and Platt]{huang2024fourier}
H.~Huang, O.~Howell, X.~Zhu, D.~Wang, R.~Walters, and R.~Platt.
\newblock Fourier transporter: Bi-equivariant robotic manipulation in 3d.
\newblock \emph{arXiv preprint arXiv:2401.12046}, 2024.

\bibitem[Pan et~al.(2023)Pan, Okorn, Zhang, Eisner, and Held]{pan2023tax}
C.~Pan, B.~Okorn, H.~Zhang, B.~Eisner, and D.~Held.
\newblock Tax-pose: Task-specific cross-pose estimation for robot manipulation.
\newblock In \emph{Conference on Robot Learning}, pages 1783--1792. PMLR, 2023.

\bibitem[van~der Pol et~al.(2021)van~der Pol, Worrall, van Hoof, Oliehoek, and Welling]{vanderpol2021mdp}
E.~van~der Pol, D.~E. Worrall, H.~van Hoof, F.~A. Oliehoek, and M.~Welling.
\newblock Mdp homomorphic networks: Group symmetries in reinforcement learning, 2021.

\bibitem[Mondal et~al.(2020)Mondal, Nair, and Siddiqi]{mondal2020group}
A.~K. Mondal, P.~Nair, and K.~Siddiqi.
\newblock Group equivariant deep reinforcement learning.
\newblock \emph{arXiv preprint arXiv:2007.03437}, 2020.

\bibitem[Fujimoto and Gu(2021)]{fujimoto2021minimalist}
S.~Fujimoto and S.~S. Gu.
\newblock A minimalist approach to offline reinforcement learning.
\newblock \emph{Advances in neural information processing systems}, 34:\penalty0 20132--20145, 2021.

\bibitem[Chebotar et~al.(2023)Chebotar, Vuong, Hausman, Xia, Lu, Irpan, Kumar, Yu, Herzog, Pertsch, et~al.]{chebotar2023q}
Y.~Chebotar, Q.~Vuong, K.~Hausman, F.~Xia, Y.~Lu, A.~Irpan, A.~Kumar, T.~Yu, A.~Herzog, K.~Pertsch, et~al.
\newblock Q-transformer: Scalable offline reinforcement learning via autoregressive q-functions.
\newblock In \emph{Conference on Robot Learning}, pages 3909--3928. PMLR, 2023.

\bibitem[Haarnoja et~al.(2018)Haarnoja, Zhou, Abbeel, and Levine]{haarnoja18soft}
T.~Haarnoja, A.~Zhou, P.~Abbeel, and S.~Levine.
\newblock Soft actor-critic: Off-policy maximum entropy deep reinforcement learning with a stochastic actor.
\newblock In J.~G. Dy and A.~Krause, editors, \emph{Proceedings of the 35th International Conference on Machine Learning, {ICML} 2018, Stockholmsm{\"{a}}ssan, Stockholm, Sweden, July 10-15, 2018}, volume~80 of \emph{Proceedings of Machine Learning Research}, pages 1856--1865. {PMLR}, 2018.

\bibitem[Wang et~al.(2022)Wang, Kohler, Zhu, Jia, and Platt]{wang2022bulletarm}
D.~Wang, C.~Kohler, X.~Zhu, M.~Jia, and R.~Platt.
\newblock Bulletarm: An open-source robotic manipulation benchmark and learning framework, 2022.

\bibitem[Mandlekar et~al.(2021)Mandlekar, Xu, Wong, Nasiriany, Wang, Kulkarni, Fei-Fei, Savarese, Zhu, and Mart\'{i}n-Mart\'{i}n]{robomimic2021}
A.~Mandlekar, D.~Xu, J.~Wong, S.~Nasiriany, C.~Wang, R.~Kulkarni, L.~Fei-Fei, S.~Savarese, Y.~Zhu, and R.~Mart\'{i}n-Mart\'{i}n.
\newblock What matters in learning from offline human demonstrations for robot manipulation.
\newblock In \emph{arXiv preprint arXiv:2108.03298}, 2021.

\bibitem[Wang et~al.(2022)Wang, Park, Sortur, Wong, Walters, and Platt]{wang2022surprising}
D.~Wang, J.~Y. Park, N.~Sortur, L.~L. Wong, R.~Walters, and R.~Platt.
\newblock The surprising effectiveness of equivariant models in domains with latent symmetry.
\newblock \emph{arXiv preprint arXiv:2211.09231}, 2022.

\bibitem[Achiam et~al.(2017)Achiam, Held, Tamar, and Abbeel]{achiam2017constrained}
J.~Achiam, D.~Held, A.~Tamar, and P.~Abbeel.
\newblock Constrained policy optimization.
\newblock In \emph{International conference on machine learning}, pages 22--31. PMLR, 2017.

\end{thebibliography}

\newpage
\appendix

\section{Notation and Preliminaries}\label{Appendix:Section:Notation and Preliminaries}

We establish some notation and review some elements of group theory.

\subsubsection{Group Theory}\label{Section:Group Theory}

At a high level, a group is the mathematical description of a symmetry. Formally, a group $G$ is a non-empty set combined with a associative binary operation $\cdot : G \times G \rightarrow G$ that satisfies the following properties
\begin{align*}
& \text{existence of identity: } e \in G, \text{ s.t. } \forall g\in G, \enspace e \cdot g = g \cdot e = g  \\
& \text{existence of inverse: } \forall g \in G, \exists g^{-1} \in G, \enspace g \cdot g^{-1} = g^{-1} \cdot g = e  \\
\end{align*}
The identity element of any group $G$ will be denoted as $e$. Note that the set consisting of just the identity element $e$ is a group.

In this study, we focus on the cyclic group of order $n$ is one of the simplest groups. $C_{n}$ is defined as the set of rotations by angle $\frac{2\pi}{n}$. Formally, the group $C_{n}$ is represented as
\begin{align*}
C_n = \{ \text{Rot}_{\theta}: \theta \in \{ \frac{2\pi i}{n} | i\in\mathbb{Z}, 0 \leq i < n \} \}
\end{align*}
Note that elements of $C_{n}$ satisfy the relation
\begin{align*}
\forall \theta,\theta', \quad \text{Rot}_{\theta} \cdot \text{Rot}_{\theta'} = \text{Rot}_{\theta + \theta'}
\end{align*}
Furthermore, the inverse of $\text{Rot}_{\theta} \in C_{n}$ is given by $\text{Rot}_{-\theta}$ which is always an element of $\text{Rot}_{-\theta} \in C_{n}$. Thus, the set $C_{n}$ is indeed a group. The total number of elements of the group $C_{n}$ is $|C_{n}| = n$.

\subsubsection{Group Actions}\label{Section:Group Actions}

Let $\Omega$ be a set. A group action $\Phi$ of $G$ on $\Omega$ is a map $\Phi : G \times \Omega \rightarrow \Omega$ which satisfies 
\begin{align}\label{Equation:Group_Action}
    &\text{Identity: } \forall \omega \in \Omega, \quad \Phi(e , \omega )  =  \omega \\
    &\nonumber \text{Compositional Property: }\forall g_{1},g_{2} \in G,\enspace \forall \omega \in \Omega, \quad  \Phi( g_{1}g_{2} , \omega  ) = \Phi(g_{1} , \Phi(g_{2}, \omega ))
\end{align}
We will often suppress the $\Phi$ function and write $\Phi( g , \omega ) = g \cdot \omega$.

\begin{center}\label{Diagram:G-Equivarient_Map}
    \begin{tikzcd}\centering
        &\Omega \arrow{d}{\Phi(g,\cdot)} \arrow{r}{ \Psi } & \Omega' \arrow{d}{ \Phi'(g, \cdot ) }  \\
        & \Omega \arrow{r}{\Psi }  & \Omega'
    \end{tikzcd}
    \captionof{figure}{Commutative Diagram For $G$-equivariant function: Let $\Phi(g, \cdot ): G \times \Omega \rightarrow \Omega$ denote the action of $G$ on $\Omega$. Let $\Phi'(g, \cdot ): G \times \Omega' \rightarrow \Omega'$ denote the action of $G$ on $\Omega'$. The map $\Psi: \Omega \rightarrow \Omega'$ is $G$-equivariant if and only if the following diagram is commutative for all $g\in G$.    }
\end{center}

Let $G$ have group action $\Phi$ on $\Omega$ and group action $\Phi'$ on $\Omega'$. A mapping $\Psi : \Omega \rightarrow \Omega'$ is said to be $G$-equivariant if and only if
\begin{align}\label{Equation:G_Equivariece_Def}
\forall g \in G, \forall \omega \in \Omega, \quad	 \Psi(  \Phi( g , \omega )  ) = \Phi'( g , \Psi(\omega) )
\end{align}
Diagrammatically, $\Psi$ is $G$-equivariant if and only if the diagram \ref{Diagram:G-Equivarient_Map} is commutative. 

\subsubsection{Group Actions of $C_n$ Group}\label{Section:Group Actions of C_n Group}

Data in two-dimensional robotic manipulation problems has natural $C_{n}$ group action. We describe three ways in which elements of $C_n$ group act on data:
\begin{enumerate}
\item $\mathbb{R}$ through trivial representation $\rho_0$: For $x \in \mathbb{R}$, the $C_n$ group acts on $x$ in the following manner, $\rho_0(g)x = x \ \forall g \in C_n$ i.e. no there is transformation under rotations. This is an example of a $C_{n}$-invariant transformation.
\item $\mathbb{R}^2$ through standard representation $\rho_1$: For $v \in \mathbb{R}^2$, the $C_n$ group acts on $v$ in the following manner, 
\begin{align*}
\forall g \in C_n \quad \rho_1(g)v =  \begin{bmatrix} cos\ g & -sin\ g \\ sin\ g  & cos\ g \end{bmatrix} v \ 
\end{align*}
Note that this is a valid group action as $\rho_{1}(g)\rho_{1}(g') v = \rho_{1}(gg') v$ for all $v \in \mathbb{R}^{2}$.
\item $\mathbb{R}^n$ through standard representation $\rho_{reg}$: For $x = (x_1,x_2,...,x_n) \in \mathbb{R}^n$, $g = r^m =C_n = \{1,r,r^2,r^3,...,r^m\}$ acts on $x$ in the following manner, $\rho_{reg}(g)x = (x_{n-m+1},...,x_n,x_1,x_2,...,x_{n-m})$ i.e. cyclically permutes the elements of $x$.
\end{enumerate}

\textbf{$C_n$ actions on feature-maps and vectors}: In general, a group element $g \in C_n$ acts on an $m$-dimensional feature map $F: \mathbb{R}^2 \rightarrow \mathbb{R}^m$ in the following manner:
\begin{align*}
(gF)(x,y) = \rho_j(g)F(\rho_1(g)^{-1}(x,y))
\end{align*}
which translates to $\rho_1 (g)$ rotating the pixel locations, and $\rho_j (g)$ transforming the pixel feature-vector using trivial representation ($\rho_j = \rho_0$), standard representation ($\rho_j = \rho_1$), or the regular representation ($\rho_j = \rho_{reg}$) depending on the nature of feature-map.

\section{Network Architectures}
Our equivariant models are implemented using the E2CNN library\cite{weiler2019general} with Pytorch. For the actor, the input is a 2-channel $\rho_0$ feature map $f_s$, and outputs a $1 \times 1$ mixed-representation type feature map representing the action, where the mixed-representation feature-map consists of 1 $\rho_1$ feature for $a_{xy}$, and 3 $\rho_0$ features for $a_{inv}=\{a_z,a_{\theta},a_{\lambda}\}$. Also, since CQL outputs the standard deviation for the conditional isotropic Gaussian distribution over the action space, the final output mixed-representation consists of  1 $\rho_1$ feature and 9 $\rho_0$ features, with additional 5 $\rho_0$ features for $a_{\sigma}$. The network consists of 8 steerable convolutional layers. 

For the critic (state-action value function), we implement a 9-layer network, where the input is a 2-channel $\rho_0$ feature map $f_s$, and a $1 \times 1$ mixed-representation type feature map representing the action, which consists  1 $\rho_1$ feature for $a_{xy}$, and 3 $\rho_0$ features for $a_{inv}$. An equivariant encoder $e$ encodes the state $f_s$ into a 64-channel $1 \times 1$ regular-representation feature-map, which is then concatenated with the action. This concatenation is then mapped to a 1-channel $1 \times 1$ trivial-representation feature-map, which represents the Q-value for the $(s,a)$. The critic also has a non-linear maxpooling layer to map intermediate regular representations to trivial representations. 

The state-value function uses a very similar architecture to the critic, except the input only includes a 2-channel $\rho_0$ feature map $f_s$, and we directly map the regular-representations to a 1-channel $1 \times 1$ trivial feature-map representing state-value, without a non-linear maxpooling layer.

The architecture of our non-equivariant models use same structure with adjusted channel dim to arrive at similar number of trainable parameters, as the equivariant networks. For more details, one can refer to the appendix for \cite{wang22so2}.

\section{Additional Experimental Results}
\begin{table}[H]
    \centering
    \small
    \centerline{
    \begin{tabular}{lc|cccc}
        \toprule
        \textbf{Task} & \textbf{Num. Demos} &  \textbf{Equi CQL} &  \textbf{Equi IQL} &  \textbf{Non-Equi CQL} & \textbf{Non-Equi IQL}\\
        \midrule
        \multirow{2}{*}{Block-in-Bowl} & 10 & \textbf{0.3181\scriptsize{$\pm$0.0392}} & \textbf{0.2871\scriptsize{$\pm$0.0559}} & 0.0025\scriptsize{$\pm$0.0036} & 0.2064\scriptsize{$\pm$0.0290}\\
        & 25 & 0.4074\scriptsize{$\pm$0.0374} & \textbf{0.5958\scriptsize{$\pm$0.0248}} &0.0\scriptsize{$\pm$0.0}& 0.4074\scriptsize{$\pm$0.0374} \\
        \midrule
        \multirow{3}{*}{Block-Stacking} & 10 & \textbf{0.1109\scriptsize{$\pm$0.0429}} & \textbf{0.1064\scriptsize{$\pm$0.0294}} & 0.0\scriptsize{$\pm$0.0}& \textbf{0.0872\scriptsize{$\pm$0.0176}} \\
        & 25 & \textbf{0.2951\scriptsize{$\pm$0.0524}} & \textbf{0.3008\scriptsize{$\pm$0.0436}} &0.0\scriptsize{$\pm$0.0} &0.0254\scriptsize{$\pm$0.0204} \\
        \bottomrule
    \end{tabular}
    }
    \caption{Best Evaluation Discounted Returns on Optimal Side View Datasets, averaged over 3 seeds (Mean and Std Deviation)}
    \label{table:SideViewOptimalDatasets}
\end{table}
In this supplementary set of experiments, we assess the performance of our equivariant methods under conditions where the task exhibits latent $SO(2)$-symmetry, yet this symmetry is not explicitly manifested in the state space. Instead of utilizing a top-down depth-image, the state is depicted through a skewed image, thereby challenging the model to discern and leverage the underlying symmetry, which is commonly scene in many robotic-manipulation settings. We use a similar setup to \cite{wang2022surprising}, where a sideview RGB-D image of the scene is used to represent the state, and a $D_4$-equivariant actor, and a $D_4$-invariant critic are learned. The $D_4$ group comprises of reflections and all the rotations of the $C_4$ group. We omit the random rotation data-augmentation used in our previous experiments as it is unsuitable for learning the $SO(2)$-latent symmetry in the sideview scenario. We evaluate our methods on optimal datasets with 10 and 25 expert demonstrations. Table \ref{table:SideViewOptimalDatasets} contains the results for the same.

Our results demonstrate that Equi-CQL and Equi-IQL outperform their non-equivariant counterparts, even when the symmetry captured by the equivariant networks doesn't exactly exist in the input image-space. These results suggest that our equivariant algorithms can be effectively applied across a diverse range of robotic applications.


Furthermore, we also compare our equivariant methods against a baseline where the non-equivariant methods are trained on a batch size that is 8-times larger than the batch-size used for training the equivariant agents. The larger batch is created by augmenting each transition in the original batch by the rotation angles of the $C_8$ group, thereby creating 8 transitions from a single transition. We train the agents on the sub-optimal datasets with 1000 transitions. The results for this experiment are reported in Table \ref{table: SubOptimalDatasets_1000_8x}

\begin{table}[t]
    \centering
    \small
    \renewcommand{\arraystretch}{1.5} 
    \centerline{
    \begin{tabular}{ll|ccccc}
        \toprule
        \textbf{Task} & \textbf{Dataset Type} &  \textbf{Equi CQL} &  \textbf{Equi IQL} &  \textbf{Non-Equi CQL} & \textbf{Non-Equi IQL}\\
        \toprule
        \multirow{2}{*}{Drawer-Opening} & Near-Random &\textbf{0.4943\scriptsize{$\pm$0.1313}}&0.2406\scriptsize{$\pm$0.2004}&0.2715\scriptsize{$\pm$0.0805}& 0.2849\scriptsize{$\pm$0.0936}\\ 
        & Medium &0.5572\scriptsize{$\pm$0.0348} &0.5903\scriptsize{$\pm$0.0045}&0.3942\scriptsize{$\pm$0.0392}& \textbf{0.6245\scriptsize{$\pm$0.0566}} \\
        \midrule
        \multirow{2}{*}{Block-in-Bowl} & Near-Random &\textbf{0.5347\scriptsize{$\pm$0.0123}}&0.3711\scriptsize{$\pm$0.0934}&0.3567\scriptsize{$\pm$0.0031}&0.3475\scriptsize{$\pm$0.1356}\\ 
        & Medium &0.5806\scriptsize{$\pm$0.0364}&\textbf{0.6305\scriptsize{$\pm$0.0456}}&0.5143\scriptsize{$\pm$0.0446} & \textbf{0.6266\scriptsize{$\pm$0.0346}} \\
        \midrule
        \multirow{2}{*}{Block-Stacking} & Near-Random &0.3086\scriptsize{$\pm$0.0143}& \textbf{0.3489\scriptsize{$\pm$0.0012}}& 0.2189\scriptsize{$\pm$0.0906}&0.2858\scriptsize{$\pm$	0.0032}\\ 
        & Medium&0.4465\scriptsize{$\pm$0.0958}& \textbf{0.4986\scriptsize{$\pm$0.0846}} &0.3897\scriptsize{$\pm$0.0665}& \textbf{0.5118\scriptsize{$\pm$0.0654}} \\
        \bottomrule
    \end{tabular}
    }
\vspace{1pt}
\caption{Best Evaluation Discounted Returns on Sub-Optimal Datasets with 1000 transitions (with augmented batches), averaged over 2 seeds. }
\label{table: SubOptimalDatasets_1000_8x}
\end{table}
We observe that Equi-CQL and Equi-IQL continue to outperform their non-equivariant baselines, even when the latter are trained on larger augmented batch sizes. Notably, there is a significant improvement in the performance of Non-Equi CQL.

\section{Analysis of the Invariant Critic} 
\begin{figure}[H]
    \centering
    \subfigure[Q-values for Invariant Critic]{%
        \includegraphics[width=0.4\textwidth]{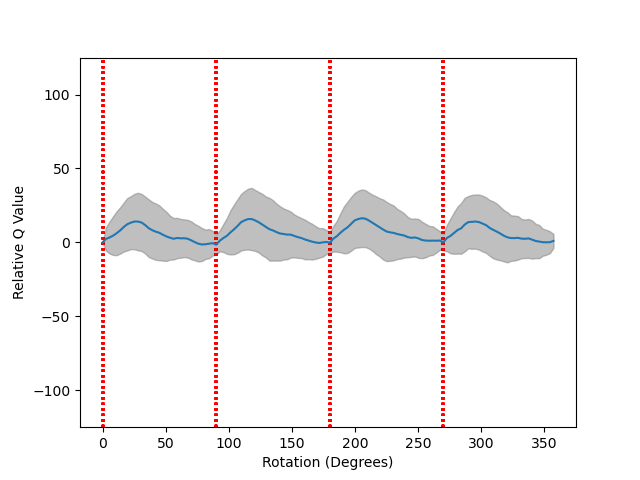}
        \label{fig:image1}
    }
    \subfigure[Q-values for Non-Invariant Critic]{%
        \includegraphics[width=0.4\textwidth]{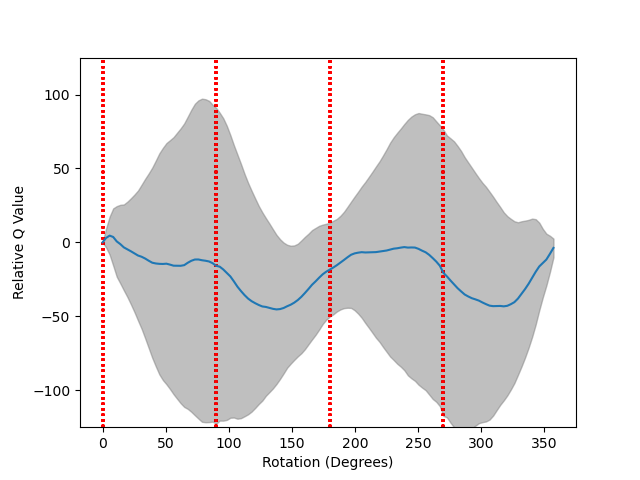}
        \label{fig:image2}
    }
    \caption{\textbf{Illustration of Normalized Learned Q-values}: The learned Q-values of the non-invariant critic are either significantly overestimated or underestimated, which can be inferred by the large standard-deviations. Furthermore, they also fail to assign consistent Q-values to rotated state-action pairs, despite the $SO(2)-$invariant nature of the task. The invariant critic assigns consistent values to rotated $(s,a)$ when the rotation-angle is a multiple of 90 degrees. Furthermore, the overestimation is also significantly smaller}
    \label{fig:Q_Values}
\end{figure}

To highlight the benefits of employing an invariant critic in training a policy using the CQL algorithm for an $SO(2)$-invariant MDP, we train a $C_4$-equivariant agent, incorporating both a $C_4$-equivariant actor and a $C_4$-invariant critic, on the sub-optimal medium dataset with 5 episodes for the block-in-bowl task. We compare this against a non-equivariant CQL agent, which is trained without the random $SO(2)$ augmentations, to underscore the performance advantages of using an $SO(2)$-invariant critic. The Equi-CQL agent outperforms the Non-Equi CQL agent by achieving a discounted return of $0.5312$, compared to Non-Equi CQL's $0.3825$. As demonstrated in our results, this performance boost can be attributed to both, the equivariance of the actor, as well as the invariance of the critic.

 The invariant critic ensures that Q-values for unseen actions are minimized, while higher Q-values are assigned to unseen optimal actions in rotated states. This capability allows for a significant deviation from the behavioral policy, a feature that Non-Equi CQL lacks. To assess this, we query both the invariant and non-invariant critics using rotated versions of the $(s,a)$ pairs from the dataset. This is achieved by uniformly sampling 128 angles within the range $[0,2\pi]$ and rotating each $(s,a)$ pair accordingly, thereby creating 128 new $(s,a)$ pairs for each $(s,a)$. We can expect some of these $(s,a)$ pair to have out-of-distribution actions that are optimal for that rotated state due to the $SO(2)$-invariant nature of our task. 

Figure \ref{fig:Q_Values} illustrates the Q-values learned by both the invariant and non-invariant critics. To represent the final Q-value for a given $(s,a)$ pair, we compute the average of the two Q-values obtained from the critic. Additionally, to account for the significant variation in Q-values among different original $(s,a)$ pairs in the dataset, we normalize the Q-values of the rotated $(s,a)$ pairs by subtracting the Q-value of the original $(s,a)$ pair. We plot the mean normalized Q-values for each rotation angle over all the $(s,a)$ pairs in the dataset and shade the region representing one standard deviation around the mean.

As seen in Fig. \ref{fig:Q_Values}, the non-invariant critic either grossly overestimates or underestimates the Q-values of rotated $(s,a)$ pairs, which are essentially comprised of out-of-distribution actions. This phenomenon is less-pronounced in the case of the invariant critic, which indicates better minimization of Q-values. Furthermore, the invariant critic assigns consistent Q-values to the rotated versions of a $(s,a)$ pair when the rotation angles are multiples of 90 degrees, due to its $C4$-invariance.This consistency can facilitate deviation from the behavioral policy, particularly when the policy is suboptimal in rotated states. In contrast, the non-invariant critic does not exhibit such invariance and can significantly underestimate the Q-values of rotated $(s,a)$ pairs. This is evident from the large standard deviations observed at 90 and 270-degree rotations.

\section{Improvement Guarantees of Equivariant-CQL}
Similar to the work of \cite{kumar20conservative}, we can show that adding symmetry bias improves the performance of the CQL algorithm. We follow a similar procedure as \cite{kumar20conservative} in our proof.

\textbf{Discounted Future-State Probability} can be defined as the probability of seeing a state $s$ in the future, given a policy $\pi$ is used to take actions. It is formally defined as:
\begin{align*}
d^\pi(s) = (1- \gamma)\sum_{t=1}^{\infty} \gamma^t P(s_t=s | \pi)
\end{align*}
Let $p^{t}_{\pi} \in  \mathbb{R}^{|S|}$  denote the vector with components $p^{t}_{\pi}(s) = P(s_t =s | \pi)$ i.e. the probability of seeing state $s$ if policy $\pi$ is used, and let $P_\pi \in \mathbb{R}^{|S|\times|S|}$ denote the transition matrix with components $p_{\pi}(s'|s) = \int da \text{ } P(s'|s, a) \pi(a|s)$ i.e. the probability of seeing next state $s'$ from current state $s$ if policy $\pi$ is used. then:
\begin{align*}
p_{t}^{\pi} = P_{\pi} p_{t-1}^{\pi} = (P_{\pi})^t \mu 
\end{align*}
\begin{align*}
d^\pi= (1- \gamma)\sum_{t=1}^{\infty} (\gamma P_\pi)^t \mu = (1- \gamma) (I - \gamma P_\pi)^{-1} \mu
\end{align*}
where $d^\pi\in  \mathbb{R}^{|S|}$ is a vector with components $d^\pi$ describing the discounted state visitation probability, and $\mu\in  \mathbb{R}^{|S|}$ is a vector with components $P(s_0 = s)$ describing the initial distribution of states.


\textbf{Invariance of Future-State Probability in Group-Invariant MDPs}
Given an equivariant policy $\pi$ ,i.e. $\pi(a|s) = \pi(ga|gs)$, in an $G-invariant$ MDP with $p_0(s)= \frac{1}{|S|} \ \forall s \in S$, we can show that the discounted state visitation probability is also $G$-invariant with $d^\pi(gs)=d^\pi(s):$
\begin{align*}
& d^\pi(s) = (1- \gamma)\sum_{t=1}^{\infty} \gamma^t P(s_t=s | \pi)\\
& d^\pi(s) = (1- \gamma)\sum_{t=1}^{\infty}\gamma^t\sum_{s_{t-1}} P(s|s_{t-1},\pi)P(s_{t-1}| \pi)
\end{align*}
We can repeat continue to expand the $P(s_{t-1}| \pi)$ till we reach $s_0$. 

For $d^\pi(gs)$ ,using the property of the $G$-invariant  MDP and a $G$-equivariant policy, $P(s' | s) = P(gs' | gs)$, we show $d^\pi(gs) = d^\pi(s)$:
\
\begin{align*}
&d^\pi(gs) = (1- \gamma)\sum_{t=1}^{\infty} \gamma^t P(s_t=gs | \pi)\\
&d^\pi(gs) = (1- \gamma)\sum_{t=1}^{\infty}\gamma^t\sum_{gs_{t-1}} P(gs|gs_{t-1},\pi)P(gs_{t-1}| \pi)\\
&d^\pi(gs) = (1- \gamma)\sum_{t=1}^{\infty}\gamma^t\sum_{s_{t-1}}P(s|s_{t-1},\pi)P(s_{t-1}| \pi)\\
&d^\pi(gs) = d^\pi(s)
\end{align*}
Thus, the discounted state visitation probability is a $G$-invariant quantity for a $G$-invariant MDP.

\textbf{A Property of Discounted Future-State Probability}: As shown in \cite{achiam2017constrained}, let there be 2 MDPs: $M$ and $\hat{M}$. Then we define the following matrices: $G = (I - \gamma P_{M}^{\pi})^{-1}$ and $G' = (I - \gamma P_{\hat{M}}^{\pi})^{-1}$ and $\Delta  = P_{\hat{M}}^{\pi} - P_{M}^{\pi}$. Using a result of \cite{kumar20conservative}, we can show that:
\begin{align*}
& d_{\hat{M}}^{\pi} - d_{M}^{\pi} = (1 - \gamma)(G' - G)\mu \\
& G^{-1} - G'^{-1} = (I - \gamma P_{M}^{\pi}) - (I - \gamma P_{\hat{M}}^{\pi})= \gamma \Delta
\end{align*}
Right-Multiplying by $G$ and Left-Multiplying by $G'$:
\begin{align*}
& G' - G = \gamma G' \Delta G\\
& d_{\hat{M}}^{\pi} - d_{M}^{\pi} = (1 - \gamma)(G' - G)\mu = (1 - \gamma)\gamma G' \Delta G\mu = \gamma G' \Delta d_{M}^{\pi}
\end{align*}
Thus, deviations between $d_{\hat{M}}^{\pi}$ and $d_{M}^{\pi}$ can be bounded by bounding the quantity $\gamma G' \Delta d_{M}^{\pi}$.

\textbf{Bounding the term $||\Delta d_{M}^{\pi}||_1$}: Similar to Lemma D.4.1 in \cite{kumar20conservative}, we want to bound $d_{M}^{\pi}$ because of its relation to $d_{\hat{M}}^{\pi} - d_{M}^{\pi}$ above. To do this we define the following bounds:
\begin{align*}
& \Delta(s' | s) = \sum_{a} (P_{\hat{M}}(s' |s,a) - P_M(s' |s,a)) \pi(a|s)\\ 
& \forall s,a \in D, |(P_{\hat{M}}(s' |s,a) - P_M(s' |s,a)| \le C/\sqrt{D(s) D(a|s)} = C/\sqrt{D(s) \pi_{\beta}(a|s)} < 1\\
& \forall s,a \notin D, |(P_{\hat{M}}(s' |s,a) - P_M(s' |s,a)| \le 1\\
\end{align*}
Here, $C$ is a constant, $D$ is the offline dataset containing transitions $\{(s,a)_i\}$, $D(s)$ is the number of times state $s$ is seen in the dataset, $D(a|s)$ is the number of times action $a$ was taken from $s$. 

Using these results, we can show that:
\begin{align*}
& ||\Delta d_{M}^{\pi}||_1 = \sum_{s'} | \sum_{s} \Delta(s' | s) d_{M}^{\pi}(s)|\\
& ||\Delta d_{M}^{\pi}||_1 \le \sum_{s',s} | \Delta(s' | s)|d_{M}^{\pi}(s) \\
& ||\Delta d_{M}^{\pi}||_1 \le \sum_{s',s}||\sum_{a}(P_{\hat{M}}(s' |s,a) - P_M(s' |s,a)) \pi(a|s)||_1 d_{M}^{\pi}(s)  \\
& ||\Delta d_{M}^{\pi}||_1 \le \sum_{s,a}||(P_{\hat{M}}(.|s,a) - P_M(.|s,a))||_1 \pi(a|s) d_{M}^{\pi}(s) \\
& ||\Delta d_{M}^{\pi}||_1 \le \sum_{s,a \in D}||(P_{\hat{M}}(.|s,a) - P_M(.|s,a))||_1 \pi(a|s) d_{M}^{\pi}(s) + \sum_{s,a \notin D}||(P_{\hat{M}}(.|s,a) - P_M(.|s,a))||_1 \pi(a|s) d_{M}^{\pi}(s) \\
& ||\Delta d_{M}^{\pi}||_1 \le \sum_{s,a \in D}\frac{C}{\sqrt{D(s) \pi_{\beta}(a|s)}} \pi(a|s) d_{M}^{\pi}(s) + \sum_{s,a \notin D}||(\pi(a|s) d_{M}^{\pi}(s) 
\end{align*}


\textbf{Equivariant Policy Learning:} An equivalent view of learning an optimal $G$-equivariant policy from an offline-dataset $D=\{(s,a,r,s')_i\}$ is to learn an optimal policy from the augmented offline dataset $D_G=\{ (gs,ga,r,gs' | g\in G, (s,a,r,s') \in D\}$. We define the empirical MDP $\hat{M}$ as the MDP formed by the transitions in the dataset $D$, and the empirical $G$-Invariant MDP as $\hat{M}_G$ formed from the transitions in $D_G$. Furthermore, we define the optimal policy learned for $\hat{M}$ as $\pi^*$, while the optimal policy learned for $\hat{M}_G$ as $\pi^*_G$. Since $\hat{M}_G$ is a $G$-invariant MDP, its optimal policy $\pi^*_G$ would be $G$-equivariant, $\pi^*_G(a | s) = \pi^*_G(ga | gs)$. Also, as $P_{\hat{M}_G}(s' | s) = P_{\hat{M}}(s' | s)$  and $P_{\hat{M}_G}(s' | gs) = 0 \ \forall g \in G \setminus \{ 0 \}, \forall (s,a,r,s') \in D$, the fixed point $\hat{Q}$ for $(s,a) \in D$ of equation (2) in the CQL paper remains the same for both the MDPs, and therefore, $\pi^*(a|s) = \pi^*_G(a|s) \ \forall (s,a,r,s') \in D$. 

 \textbf{Bounding the term $||\Delta d_{M}^{\pi}||_1$ for equivariant and non-equivariant policies:} Based on our proofs, we can show $||\Delta_{\hat{M}_G} d_{M}^{\pi^*_{G}}|| <  ||\Delta_{\hat{M}} d_{M}^{\pi^*}||$. For $||\Delta_{\hat{M}} d_{M}^{\pi^*}||_1$:
 \begin{align*}
 & ||\Delta_{\hat{M}} d_{M}^{\pi^*}||_1 \le \sum_{s,a \in D}\frac{C}{\sqrt{D(s) \pi_{\beta}(a|s)}} \pi^*(a|s) d_{M}^{\pi^*}(s) + \sum_{s,a \notin D}(\pi^*(a|s) d_{M}^{\pi^*}(s) \\ 
 \end{align*}
 
 If we define the cardinally of Group $G$ as $|G|$, then the size of the augmented Dataset $D_G$ would be $|D_G| = |G| |D|$ i.e. $|G|$ times the size of the original dataset. Furthermore, $D(s) = D(gs), \pi_{\beta}(a|s)=\pi_{\beta}(ga|gs) \forall g \in G$
 Then:
 \begin{align*}
 ||\Delta_{\hat{M}_G} d_{M}^{\pi^*_{G}}||_1 \le \sum_{s,a \in D_G}\frac{C}{\sqrt{D(s) \pi_{\beta}(a|s)}} \pi^*_{G}(a|s) d_{M}^{\pi^*_{G}}(s) + \sum_{s,a \notin D_G}(\pi^*_{G}(a|s) d_{M}^{\pi^*_{G}}(s) \\ \\
 ||\Delta_{\hat{M}_G} d_{M}^{\pi^*_{G}}||_1 \le |G|\sum_{s,a \in D}\frac{C}{\sqrt{D(s) \pi_{\beta}(a|s)}} \pi^*_{G}(a|s) d_{M}^{\pi^*_{G}}(s) + ((\sum_{s,a \notin D}(\pi^*_{G}(a|s) d_{M}^{\pi^*_{G}}(s)) - \\
 (|G| - 1)  \sum_{s,a \in D}\pi^*_{G}(a|s) d_{M}^{\pi^*_{G}}(s))\\
\end{align*}


We can show that the upper-bound for $||\Delta_{\hat{M}} d_{M}^{\pi^*}||_1$, represented as $||\Delta_{\hat{M}} d_{M}^{\pi^*}||_1^{U}$, is larger than the upper-bound of $||\Delta_{\hat{M}_G} d_{M}^{\pi^*_{G}}||_1$, represented as $||\Delta_{\hat{M}_G} d_{M}^{\pi^*_{G}}||_1^{U}$:
\begin{align*}
& ||\Delta_{\hat{M}} d_{M}^{\pi^*}||_1^U - ||\Delta_{\hat{M}_G} d_{M}^{\pi^*_{G}}||_1^U = (|G| - 1)(\sum_{s,a \in D}\pi^*_{G}(a|s) d_{M}^{\pi^*_{G}}(s)) - \sum_{s,a \in D}\frac{C}{\sqrt{D(s) \pi_{\beta}(a|s)}} \pi^*_{G}(a|s) d_{M}^{\pi^*_{G}}(s))\\
& (|G| - 1)(\sum_{s,a \in D}\pi^*_{G}(a|s) d_{M}^{\pi^*_{G}}(s)) - \sum_{s,a \in D}\frac{C}{\sqrt{D(s) \pi_{\beta}(a|s)}} \pi^*_{G}(a|s) d_{M}^{\pi^*_{G}}(s)) > 0\\ 
 & ||\Delta_{\hat{M}} d_{M}^{\pi^*}||_1^U - ||\Delta_{\hat{M}_G} d_{M}^{\pi^*_{G}}||_1^U > 0 \\ 
 & ||\Delta_{\hat{M}} d_{M}^{\pi^*}||_1^U > ||\Delta_{\hat{M}_G} d_{M}^{\pi^*_{G}}||_1^U
\end{align*}


By using our derived bounds for the term $||\Delta d_{M}^{\pi}||_1$ for the equivariant and non-equivariant policy in Lemma D.4.1 and Theorem D.4 in \cite{kumar20conservative}, the obtained upper-bound for $|J(\pi^*_{G},\hat{M}_G) - J(\pi^*_{G},M)|$ will be smaller in comparison to $|J(\pi^*,\hat{M}) - J(\pi^*,M)|$,. This, in turn, would result in a bigger lower-bound for $(J(\pi^*_{G},M) - J(\pi_{\beta},M))$ in comparison to $(J(\pi^*,M) - J(\pi_{\beta},M))$. 

From the above result, we can conclude that the worst-possible expected return of the learned equivariant policy $J(\pi^*_{G},M)$ would always be greater than the worst-possible expected return of the learned non-equivariant policy $J(\pi^*,M)$.

\end{document}